\documentclass[runningheads]{llncs}
\usepackage{graphicx}
\usepackage{multirow}
\usepackage{amsmath,amssymb} 
\usepackage{color}
\usepackage[width=122mm,left=12mm,paperwidth=146mm,height=193mm,top=12mm,paperheight=217mm]{geometry}

\begin{document}
\pagestyle{headings}
\mainmatter

\title{Saliency Detection via Combining Region-Level and Pixel-Level Predictions with CNNs} 

\titlerunning{Saliency Detection via Combining Region-Level and Pixel-Level Predictions}

\authorrunning{Youbao Tang, Xiangqian Wu}

\author{Youbao Tang, Xiangqian Wu}


\institute{Harbin Institute of Technology, Harbin, 150001, China \\ \email{\{tangyoubao, xqwu\}@hit.edu.cn}}

\maketitle

\begin{abstract}
This paper proposes a novel saliency detection method by combining region-level saliency estimation and pixel-level saliency prediction with CNNs (denoted as CRPSD). For pixel-level saliency prediction, a fully convolutional neural network (called pixel-level CNN) is constructed by modifying the VGGNet architecture to perform multi-scale feature learning, based on which an image-to-image prediction is conducted to accomplish the pixel-level saliency detection. For region-level saliency estimation, an adaptive superpixel based region generation technique is first designed to partition an image into regions, based on which the region-level saliency is estimated by using a CNN model (called region-level CNN). The pixel-level and region-level saliencies are fused to form the final salient map by using another CNN (called fusion CNN). And the pixel-level CNN and fusion CNN are jointly learned. Extensive quantitative and qualitative experiments on four public benchmark datasets demonstrate that the proposed method greatly outperforms the state-of-the-art saliency detection approaches.

\end{abstract}

\section{Introduction}
Visual saliency detection, which is an important and challenging task in computer vision, aims to highlight the most important object regions in an image. Numerous image processing applications incorporate the visual saliency to improve their performance, such as image segmentation \cite{jung2012unified} and cropping \cite{rother2006autocollage}, object detection \cite{luo2014switchable}, and image retrieval \cite{gao2013visual}, etc.

The main task of saliency detection is to extract discriminative features to represent the properties of pixels or regions and use machine learning algorithms to compute salient scores to measure their importances. A large number of saliency detection approaches \cite{5rosin2009simple,6liu2011learning,7yan2013hierarchical,9yang2013saliency,10li2013saliency,11fi2013salient,12cheng2013efficient,13zhang2013saliency,14jiang2013saliency,15li2013contextual,16liu2014adaptive,17lu2014learning,18so2014saliency,19kim2014salient,20li2014secrets,49tang2015saliency,21li2015visual,22frintrop2015traditional,23tong2015salient,24gong2015saliency,25qin2015saliency,26zhao2015saliency,27li2015robust,28wang2015deep,29li2015weighted,30zhang2015minimum,31wang2015pisa,32li2015inner,33sun2015saliency,34cheng2015pami,35scharfenberger2015structure,36achanta2009frequency} have been proposed by exploiting different salient cues recently. They can be roughly categorized as pixel based approaches and region based approaches. For the pixel based approaches, the local and global features, including edges \cite{5rosin2009simple}, color difference \cite{36achanta2009frequency}, spatial information \cite{6liu2011learning}, distance transformation \cite{30zhang2015minimum}, and so on, are extracted from pixels for saliency detection. Generally, these approaches highlight high contrast edges instead of the salient objects, or get low contrast salient maps. That is because the extracted features are unable to capture the high-level and multi-scale information of pixels. As we know that convolutional neural network (CNN) is powerful for high-level and multi-scale feature learning and has been successfully used in many applications of computer vision, such as semantic segmentation \cite{37zheng2015conditional,38long2015fully}, edge detection \cite{39xie2015holistically,40bertasius2015deepedge}, etc. This work will employ CNN for pixel-level saliency detection.

\begin{figure}
	\centering
	\includegraphics[width=0.9\textwidth]{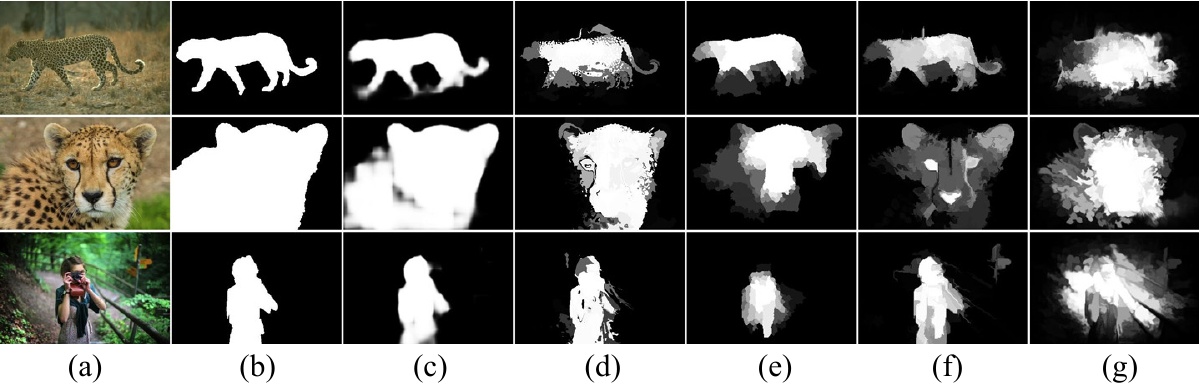}
	\caption{Three examples of saliency detection results estimated by the proposed method and the state-of-the-art approaches. (a) The input images. (b) The ground truths. (c) The salient maps detected by the proposed method. (d)-(g) The salient maps detected by the state-of-the-art approaches MC \cite{26zhao2015saliency}, MDF \cite{21li2015visual}, LEGS \cite{28wang2015deep}, and MB+ \cite{30zhang2015minimum}.}
	\label{fig1}
\end{figure}

For the region based approaches, they first segment an image into a number of regions, and then many different kinds of hand-designed features \cite{7yan2013hierarchical,9yang2013saliency,10li2013saliency,11fi2013salient,18so2014saliency,19kim2014salient,23tong2015salient,25qin2015saliency,27li2015robust,32li2015inner,33sun2015saliency,34cheng2015pami,35scharfenberger2015structure} and CNN based features \cite{21li2015visual,26zhao2015saliency,28wang2015deep} are extracted to compute the salienies from these regions. Compared with the pixel based approaches, these regions based approaches are more effective to detect the saliency since more sophisticated and discriminative features can be extracted from regions. The approaches based on CNN learned features have gotten better performance than the ones based on hand-designed features. That is because CNN is able to extract more robust and discriminative features with considering the global context information of regions. Therefore, this work also employs CNN for region-level saliency estimation. Recently, the best region based saliency detection approach proposed by Zhao et al \cite{26zhao2015saliency} extracts superpixels as regions, then estimates the saliency for each superpixel based on CNN. In their work, an inevitable problem is that it is hard to decide the number of superpixels. If there are too few superpixels, the regions belonging to salient objects may be under-segmented. If there are too many superpixels, the regions belonging to saliency objects or backgrounds may be over-segmented, which may cause that the saliencies are not uniform in salient objects or backgrounds, and the superpixels around the boundaries of background and salient objects may get wrong saliencies. Furthermore, the number of superpixels should be different according to the complexity of images. In this paper, we follow their work and propose an adaptive superpixel based region generation technique, which can automatically determine the number of generated regions for different images to solve the above-mentioned problems and improve the performance of saliency detection.

Since pixel-level and region-level saliency detection approaches make use of different information of images, these two salient maps are complementary. Hence, we propose a CNN network to fuse the pixel-level and the region-level saliencies to improve the performance. Fig. \ref{fig1} shows some results of the proposed method, which are very close to the ground truths.

\begin{figure}
	\centering
	\includegraphics[width=0.8\textwidth]{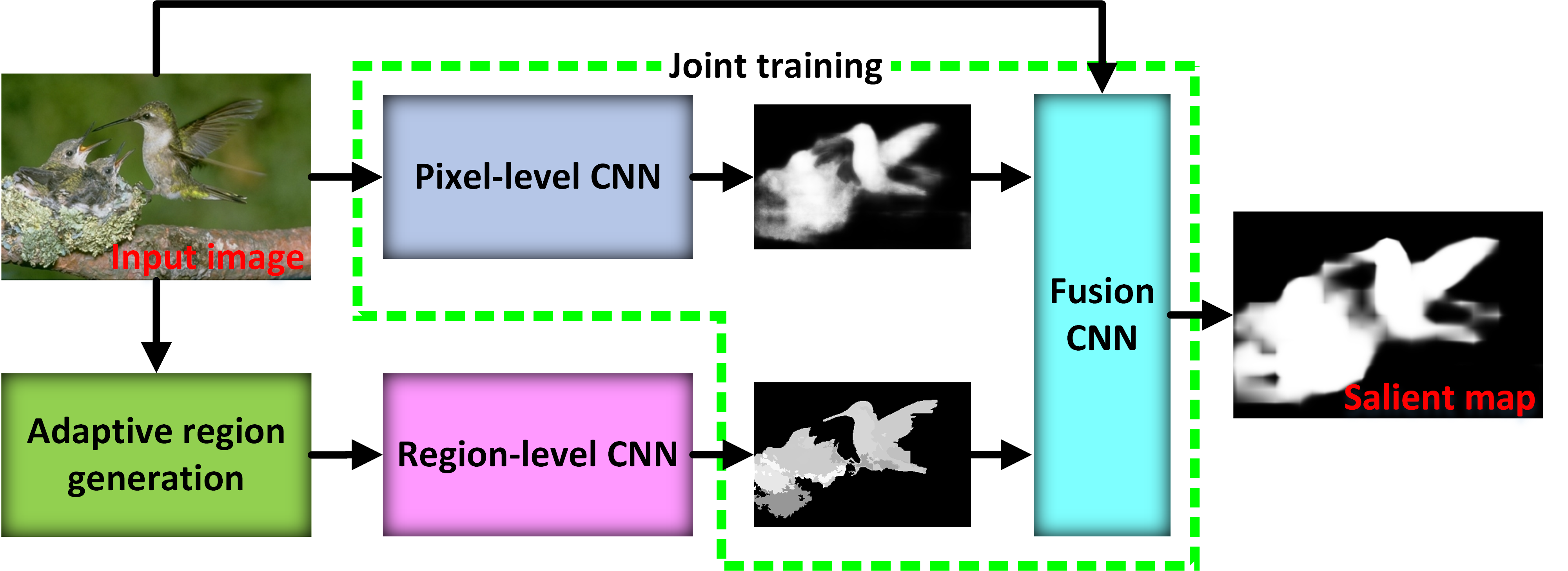}
	\caption{The framework of the proposed method.}
	\label{fig2}
\end{figure}

Fig. \ref{fig2} shows the framework of proposed method, which consists of three stages, i.e. pixel-level saliency prediction, region-level saliency estimation, and the salient map fusion. 
For pixel-level saliency prediction, a pixel-level CNN is constructed by modifying the VGGNet \cite{41vgg2014very} and finetuning from the pre-trained VGGNet model for pixel-level saliency prediction. 
For region-level saliency estimation, the input image is first segmented into a number of regions by using an adaptive superpixel based region generation technique. Then for each region, a salient score is estimated based on a region-level CNN. 
For salient map fusion, the pixel-level and region-level salient maps are fused to form the final salient map by using a fusion CNN which is jointly trained with the pixel-level CNN.

The main contributions of this paper are summarized as follows. (1) A novel multiple CNN framework is proposed to extract and combine pixel and region information of images for saliency detection. (2) A pixel-level CNN is devised for pixel-level saliency prediction. (3) An adaptive region generation technique is developed to generate regions and based on which a region-level CNN is used for region-level saliency estimation. (4) A fusion-level CNN is proposed to fuse the pixel-level and region-level saliencies.

\section{Pixel-level saliency prediction}
CNN has achieved a great success in various applications of computer vision, such as classification and segmentation. Here, we proposed a CNN (denoted as pixel-level CNN) to predict the saliency for each pixel. Pixel-level CNN takes the original image as the input and the salient map as the output. To get an accurate saliency prediction, the CNN architecture should be deep and have multi-scale stages with different strides, so as to learn discriminative and multi-scale features for pixels. Training such a deep network from scratch is difficult when the training samples is not enough. However, there are several networks which have achieved the state-of-the-art results in the ImageNet challenge, such as VGGNet \cite{41vgg2014very} and GoogleNet \cite{42googlenet2015going}. So it is an effective way to use these excellent models trained on the large-scale dataset as the pre-trained model for finetuning.

\begin{figure}
	\centering
	\includegraphics[width=0.75\textwidth]{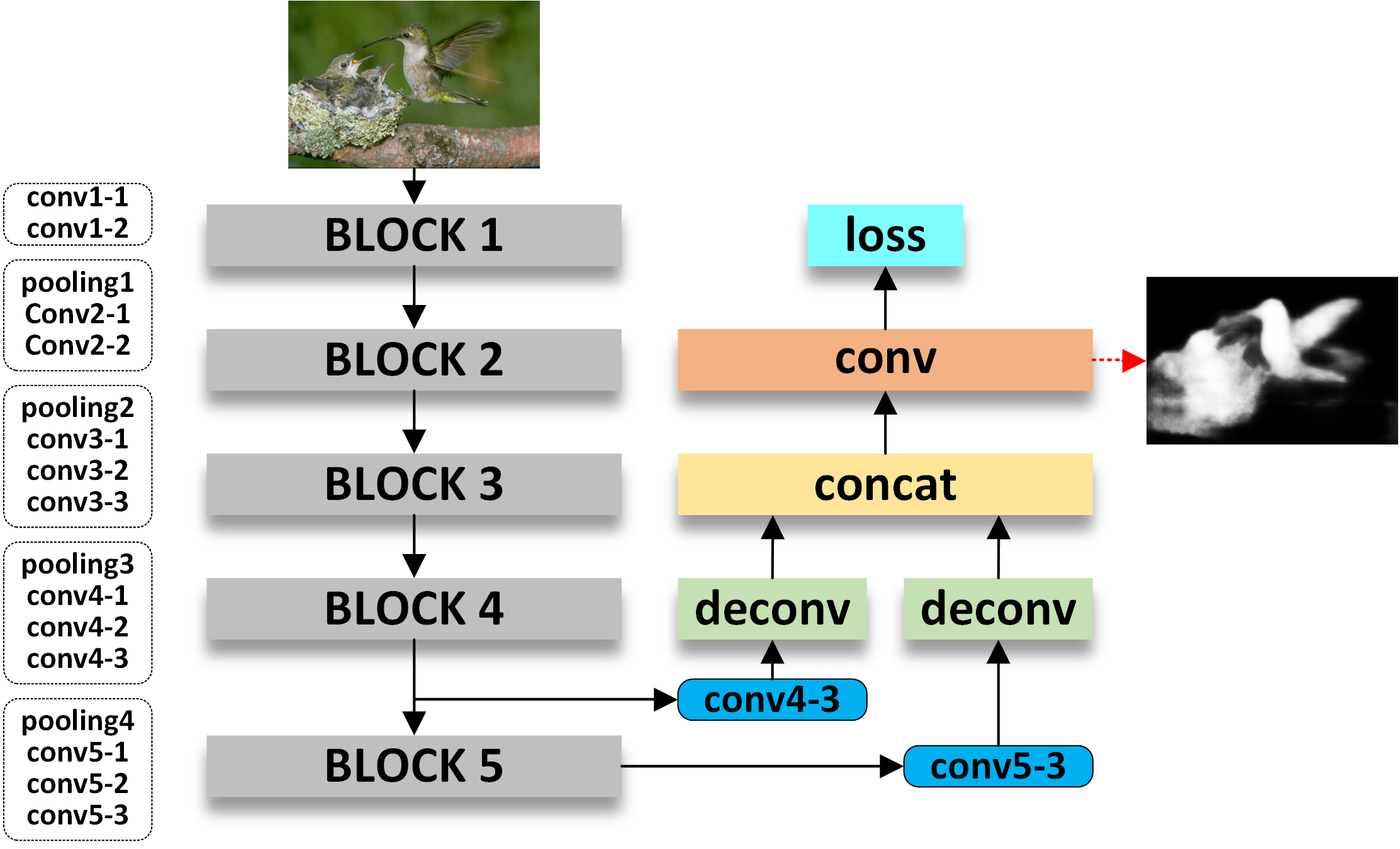}
	\caption{The architecture of the pixel-level CNN network.}
	\label{fig3}
\end{figure}

In this work, we construct a deep CNN architecture based on VGGNet for pixel-level saliency prediction. The VGGNet consists of six blocks. The first five blocks contain convolutional layers and pooling layers, as shown in Fig. \ref{fig3}. The last block contains one pooling layer and two fully connected layer, which are used to form the final feature vector for image classification. While for saliency prediction, we need to modify the VGGNet to extract dense pixel-level features. Therefore, the last block is removed in this work. There are two main reasons for this modification. The first one is that the fully connected layers cost much time and memory during training and testing. The second one is that the output of the last pooling layer is too small compared with the original image, which will reduce the accuracy of fullsize prediction. In order to capture the multi-scale information, we combine the outputs of the last two blocks of the modified VGGNet for the multi-scale feature learning. The benefits of doing such combination is two-fold. The first one is that the receptive field size becomes larger when the output size of blocks becomes smaller. Therefore, the output combination of multiple blocks can automatically learn the multi-scale features. The second one is that the shallow blocks mainly learn the local features, such as edges and parts of objects, which are not very useful for saliency detection since we hope to capture the global information of whole salient objects. Therefore, the outputs of the last two blocks are combined for multi-scale feature learning.

\begin{figure}
	\centering
	\includegraphics[width=0.75\textwidth]{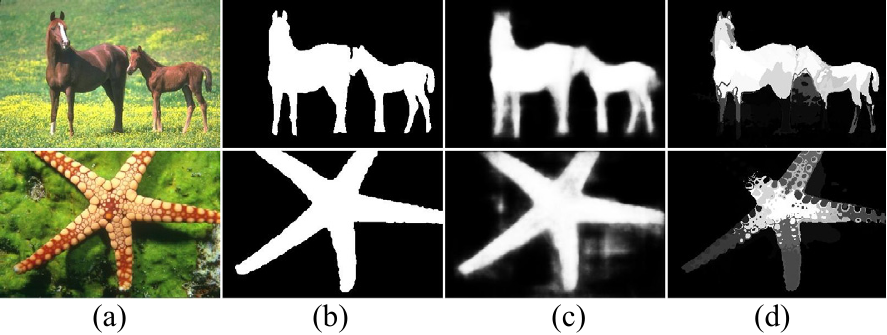}
	\caption{Examples of pixel-level saliency prediction results. (a) Original images. (b) Ground truths. (c) Pixel-level saliency prediction results. (d) Salient maps estimated by the state-of-the-art approach MC \cite{26zhao2015saliency}.}
	\label{fig4}
\end{figure}

Since the output sizes of the last two blocks are different and smaller than the size of the input image. To make the whole CNN network automatically learn the multi-scale features for pixel-level saliency prediction, we first perform the deconvolutional operation for the outputs of the last two blocks to make them have the same size with the input image, and concatenate them in the channel direction. Then a convolutional kernel with size of $ 1\times 1$ is used to map the concatenation feature maps into a probability map, in which larger values mean more saliencies. For testing, the probability map actually is a salient map of the input image. For training, a loss function is needed to compute the errors between the probability map and the ground truth. For most of the images, the numbers of salient and non-salient pixels are heavily imbalanced. Therefore, given an image $ X $ and its ground truth $ Y $, a cross-entropy loss function is used to balance the loss between salient and non-salient classes as follows:
\begin{align}
	L\left( \textbf{W} \right) = -\alpha\sum_{i=1}^{\vert Y_+ \vert}\log P\left( y_i=1 \vert X, \textbf{W} \right) -\left( 1-\alpha\right)\sum_{i=1}^{\vert Y_- \vert}\log P\left( y_i=0 \vert X, \textbf{W} \right)
\end{align}
where $\alpha=\vert Y_- \vert / \left(\vert Y_+ \vert+\vert Y_- \vert \right)$, $\vert Y_+ \vert$ and $\vert Y_- \vert$ mean the number of salient pixels and non-salient pixels in ground truth, and $\textbf{W}$ denotes the parameters of all network layers. Here and now, the whole pixel-level CNN architecture is constructed as shown in Fig. \ref{fig3}. The standard stochastic gradient descent algorithm is used to minimize the above loss function during training. After training, given an image, we can use the trained CNN model to predict a pixel-level salient map. Fig. \ref{fig4} shows two examples of pixel-level saliency prediction resultss.

\section{Region-level saliency estimation}
Inspired by the successful application of CNN in salient object detection \cite{21li2015visual,26zhao2015saliency,28wang2015deep}, all of which are based on regions (e.g. superpixels \cite{26zhao2015saliency} and multi-scale regions \cite{21li2015visual}), this work also employs CNN for the region-level saliency estimation.

\subsection{Adaptive region generation}
During the region-level saliency estimation, the first step is to generate a number of regions from the input image. Wang et al \cite{28wang2015deep} use the regions in sliding windows to estimate their saliencies, which may result in the salient object and background in the same sliding window having the same saliency. Li et al \cite{21li2015visual} use multi-scale hierarchical regions, which consumes much time to perform the region segmentation and some generated regions are under-segmented. Zhao et al \cite{26zhao2015saliency} use superpixels as the regions to estimate their saliencies, which is difficult to decide the number of superpixels. If there are too few superpixels, the regions belonging to salient objects may be under-segmented. If there are too many superpixels, the regions belonging to saliency objects or backgrounds may be over-segmented. Both over-segmentation and under-segmentation may make the saliencies are not uniform in salient objects or backgrounds. Different images should be segmented into different number of superpixels because of their different properties.

Since the superpixels based approach \cite{26zhao2015saliency} gets the state-of-the-art performance, this work proposes an adaptive region generation technique based on this approach to segment the images and solve the abovementioned problems.

\begin{figure}
	\centering
	\includegraphics[width=0.75\textwidth]{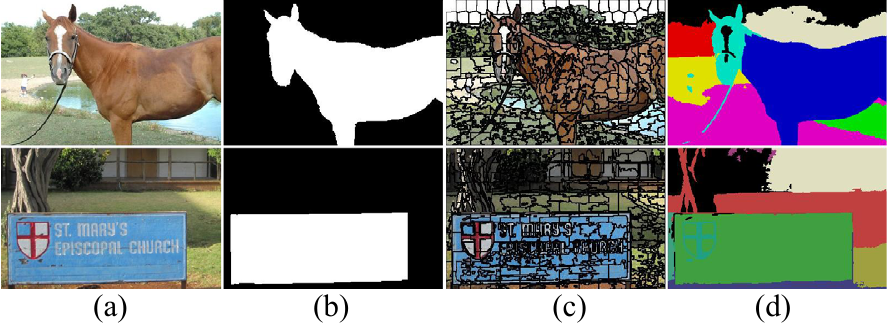}
	\caption{Examples of our adaptive region generation technique. (a) Original images. (b) Ground truths. (c) Superpixel segmentation results. (d) Region generation results.}
	\label{fig5}
\end{figure}

Given an input image $ I $, it is first over-segmented into $ n $ superpixels by using SLIC algorithm \cite{43achanta2012slic}. Here, we set $ n=300 $ with considering both of effectiveness and efficiency. Then for each superpixel, a simple feature vector including its average colors in L*a*b color space and average spatial coordinates is computed. Then a graph-based agglomerative clustering algorithm (called Graph Degree Linkage) \cite{44zhang2012graph}, which takes the superpixel as nodes and assigns each node with $ k $ edges whose weights are computed according to the Euclidean distances between the feature vectors of the current node and its $ k $ nearest neighbor nodes, is used to cluster the superpixels into different regions. The clustering process is stopped when the least affinity between two clusters is larger than a given threshold $ t $. Therefore, for different images, the numbers of clustered regions are different and are much less than $ n $. The superpixels which are adjacent and have similar colors are usually clustered into the same regions. The whole clustering process has two important parameters $ k $ and $ t $, which are set as $ k=15 $ and $ t=-0.04 $ through experiments in this work. Fig. \ref{fig5} shows two examples of region generation results.

\subsection{Region saliency estimation}
After obtaining the regions, the next step is to estimate the regions’ saliencies. This work employs CNN for region-level saliency estimation. The Clarifai model \cite{45zeiler2014visualizing}, which is the winning model in the classification task of ImageNet 2013, is used as our CNN model as done by \cite{26zhao2015saliency}. It contains five convolutional layers and two fully connected layers. For more detail information about this model, please refer to the reference \cite{45zeiler2014visualizing}. In this work, we use the CNN model provided by the authors of \cite{26zhao2015saliency} as the pre-trained model and finetune for the region-level saliency estimation.

\begin{figure}
	\centering
	\includegraphics[width=0.75\textwidth]{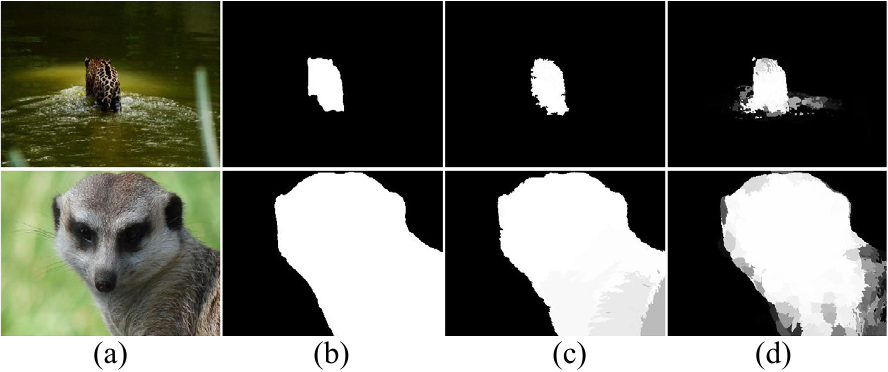}
	\caption{Examples of region-level saliency estimation results. (a) Original images. (b) Ground truths. (c) Salient maps estimated by the proposed region-level saliency estimation method. (d) Salient maps estimated by superpixel based region saliency estimation method .}
	\label{fig6}
\end{figure}

In \cite{26zhao2015saliency}, the region in a superpixel-centered large context window is resized and fed into the CNN model to estimate the saliency of current superpixel. If we follow the same way except using region-centered instead of superpixel-centered, a problem will be introduced, that is some background regions may have large saliencies, because the centers of some background regions may belong to or close to the salient objects. To solve this problem, we randomly choose $ m $ superpixels around the centerline of each region at first. Then we set these $ m $ superpixels’ centers as the windows’ centers to construct $ m $ large context windows including the full image as done by \cite{26zhao2015saliency}. We choose superpixels around the region’s centerline to make the windows’ centers far away from the regions’ boundaries as much as possible, and the constructed windows from different regions are different as much as possible. Here, we set $ m=5 $ if the number of superpixels in a region is larger than 5. Otherwise, we set $ m $ as the number of superpixels. Through experiments, we find that the performances of saliency detection vary little when $ m>5 $.

For each region, we can construct $ m $ window images and feed them into the CNN model to obtain $ m $ saliencies. In this work, the mean saliency is computed as the region’s saliency due to its robustness to noises. Compared with the superpixel-centered saliency estimation approach, the proposed region-level saliency estimation method has three advantages described as follows. (1) More efficiency, because the constructed images are much less than the superpixels. (2) Less boundary effect, which is that the salient regions around the boundaries of salient objects and backgrounds may have small saliencies while the background regions around the boundaries may have large saliencies, as shown in Fig. \ref{fig6}. (3) More uniform salient map, since the pixels in a region are assigned the same salient values, as shown in Fig. \ref{fig6}.

\section{Salient map fusion}
Given an input RGB image, the proposed saliency detection method efficiently produces two salient maps, i.e. region-level salient map and the pixel-level salient map. These two salient maps are computed by using different information of images, hence they are complementary and can be fused to further improve the performance.

\begin{figure}
	\centering
	\includegraphics[width=0.8\textwidth]{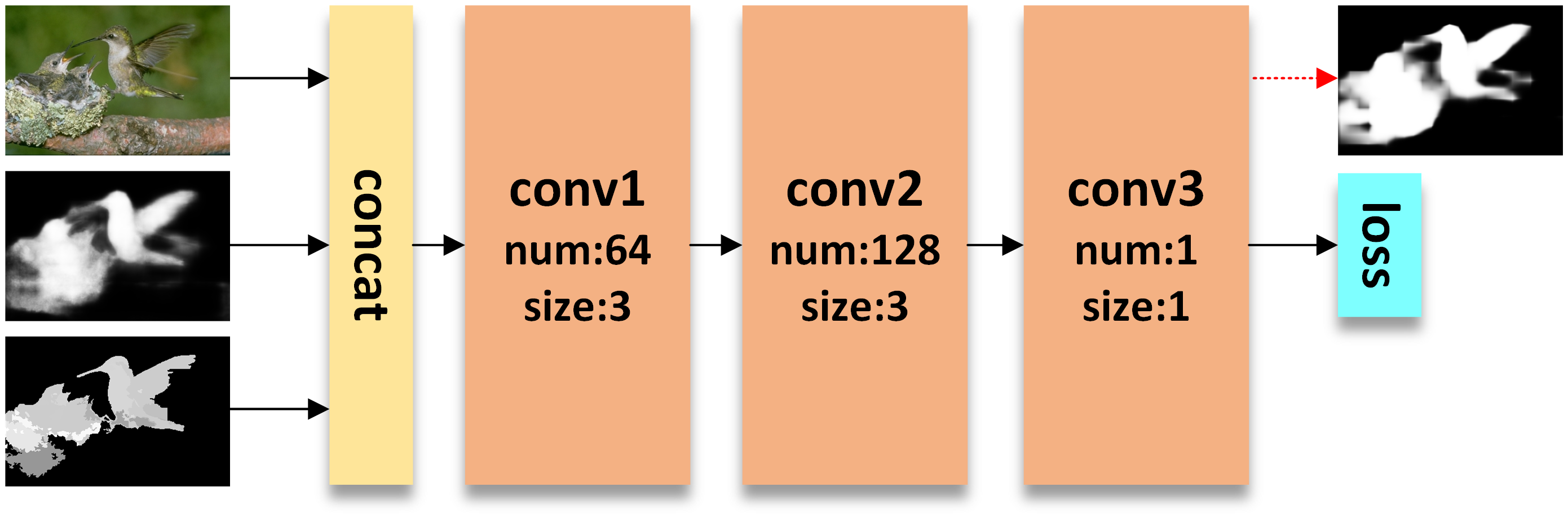}
	\caption{The architecture of the fusion CNN network.}
	\label{fig7}
\end{figure}

There are many fusion strategies, such as establishing some measures to select a better individual salient map \cite{12cheng2013efficient} or combining salient maps with weighted values \cite{7yan2013hierarchical}. They don't use the information of all salient maps or only linearly combine them. In this work, we sufficiently dig their complementary information with a nonlinear manner to improve the performance by using a CNN network. The CNN network contains one concatenation layer, three convolutional layers, and a loss layer, as shown in Fig. \ref{fig7}. The input image and its two salient maps are concatenated into a 5-channel image, and then through three convolutional layers whose configures are given in Fig. \ref{fig7}. For testing, the output of the last convolutional layer is the prediction salient map. For training, the loss layer is used to compute the errors between the output of the last convolutional layer and the ground truth with the cross-entropy loss function described before. It is needed to be noticed that the original image also is used for fusion except two salient maps. That's because richer information of original images is incorporated to correct some errors which cannot be solved by only using the salient maps.

The fusion CNN network can be trained separately. But as we know that joint training multiple sub-networks can gain the performance improvement. In this work, the region-level salient estimation needs to generate a number of regions at the begin-ning and the region-level CNN has a big different with the pixel-level CNN and fusion CNN. So it is hard to treat all of these three CNN network as an end-to-end network for joint training. Finally, the region-level CNN is trained alone, and after that, the pixel-level CNN and fusion CNN are jointly trained to get the final salient map as shown in Fig. \ref{fig2}. Based on the final salient maps, some post-processings, such as fully connected CRF \cite{koltun2011efficient}, can be used to further improve the performance. But in this work, to focus on the performance of saliency detection models, we don't conduct any post-processing.

\section{Experiments}

\subsection{Implementation}
We use the popular Caffe library \cite{46jia2014caffe} to implement the proposed saliency detection framework. The THUS-10000 dataset \cite{34cheng2015pami} contains 10,000 images and their corresponding ground truths, which is used for CNN model training. For the region-level CNN network training, we use the Clarifai model trained by \cite{26zhao2015saliency} as the pre-trained model to finetune on the training dataset. Before joint training the pixel-level CNN and fusion CNN network, we separately train them to get the initial models. For the pixel-level CNN network, since it is a fully convolutional network, arbitrary images don't need to be resized. And the weights of the first five blocks of VGGNet model trained on ImageNet are used to do the weight initialization, based on which the modified VGGNet is finetuned for pixel-level saliency prediction. For the fusion CNN network, we train the model from scratch. After obtaining the initial models of pixel-level and fusion CNN network, we use the weights of these models as weight initialization of the joint CNN network and use the training dataset to do the end-to-end training. The above training process costs about 49 hours for 30,000 iterations on a PC with an Intel i7-4790k CPU, a TESLA k40c GPU, and 32G RAM. For testing on an image with the size of $ 300\times 400 $, the region-level saliency estimation takes about 0.5 second, the process of pixel-level saliency prediction and saliency fusion takes about 0.38 second. Therefore, the whole process time of our saliency detection method is about 0.88 second.

\subsection{Datasets and evaluation criteria}
\subsubsection{Datasets.}
We evaluate the proposed method on four standard benchmark datasets: SED \cite{47alpert2012image}, ECSSD \cite{7yan2013hierarchical}, PASCAL-S \cite{20li2014secrets}, and HKU-IS \cite{21li2015visual}.

SED \cite{47alpert2012image} contains 200 images with one or two salient object, in which objects have largely different sizes and locations. This dataset is the combination of SED1 and SED2 dataset.

ECSSD \cite{7yan2013hierarchical} contains 1,000 images with complex backgrounds, which makes the detection tasks much more challenging.

PASCAL-S \cite{20li2014secrets} is constructed on the validation set of the PASCAL VOC 2012 segmentation challenge. This dataset contains 850 natural images with multiple complex objects and cluttered backgrounds. The PASCAL-S data set is arguably one of the most challenging saliency data sets without various design biases (e.g., center bias and color contrast bias).

HKU-IS \cite{21li2015visual} contains 4447 challenging images, which is newly developed by considering at least one of the following criteria: (1) there are multiple disconnected salient objects, (2) at least one of the salient objects touches the image boundary, (3) the color contrast (the minimum Chi-square distance between the color histograms of any salient object and its surrounding regions) is less than 0.7.

All datasets provide the corresponding ground truths in the form of accurate pixel-wise human-marked labels for salient regions.

\subsubsection{Evaluation criteria.}
The standard precision-recall (PR) curves are used for performance evaluation. Precision corresponds to the percentage of salient pixels correctly assigned, while recall corresponds to the fraction of detected salient pixels in relation to the ground truth number of salient pixels. The PR curves are obtained by binarizing the saliency map in the range of 0 and 255. The F-measure ($ F_\beta $) is the overall performance measurement computed by the weighted harmonic of precision and recall:
\begin{align}
	F_\beta = \dfrac{\left( 1+\beta^2\right) \times Precision \times Recall}{\beta^2 \times Precision + Recall}
\end{align}
where we set $\beta^2=0.3$, as done by other approaches.

The mean absolute error ($ MAE $), which is the average per-pixel difference between the ground truth $ GT $ and the saliency map $ S $, is also evaluated. Here, $ GT $ and $ S $ are normalized to the interval [0, 1]. $ MAE $ is defined as 
\begin{align}
	MAE = \dfrac{\sum\limits_{x=1}^W \sum\limits_{y=1}^H \vert S \left(x,y \right)-GT \left(x,y \right) \vert}{W \times H}
\end{align}
where $ W $ and $ H $ are the width and height of the image.

We also adopt the weighted $F_\beta$ metric \cite{48margolin2014evaluate} (denoted as $wF_\beta$) for evaluation, which suffers less from curve interpolation flaw, improper assumptions about the independence between pixels, and equal importance assignment to all errors. We use the code and the default setting of $wF_\beta$ provided by the authors of \cite{48margolin2014evaluate}.

\subsection{Performance comparisons with state-of-the-art approaches}
We compare the proposed method (denoted as CRPSD) and the two submodules (pixel-level saliency prediction, denoted as PSD, and region-level saliency estimation, denoted as RSD) with seventeen existing state-of-the-art saliency detection approaches on four datasets, including MC \cite{26zhao2015saliency}, MDF \cite{21li2015visual}, LEGS \cite{28wang2015deep}, CPISA \cite{31wang2015pisa}, MB+ \cite{30zhang2015minimum}, SO \cite{18so2014saliency}, BSCA \cite{25qin2015saliency}, DRFI \cite{11fi2013salient}, DSR \cite{10li2013saliency}, LPS \cite{32li2015inner}, MAP \cite{33sun2015saliency}, MR \cite{9yang2013saliency}, RC \cite{34cheng2015pami}, RRWR \cite{27li2015robust}, SGTD \cite{35scharfenberger2015structure}, BL \cite{23tong2015salient}, and HS \cite{7yan2013hierarchical}. For fair comparison, the source codes of these state-of-the-art approaches released by the authors are used for test with recommended parameter settings in this work.

\begin{figure}[!htb]
	\centering
	\includegraphics[width=0.98\textwidth]{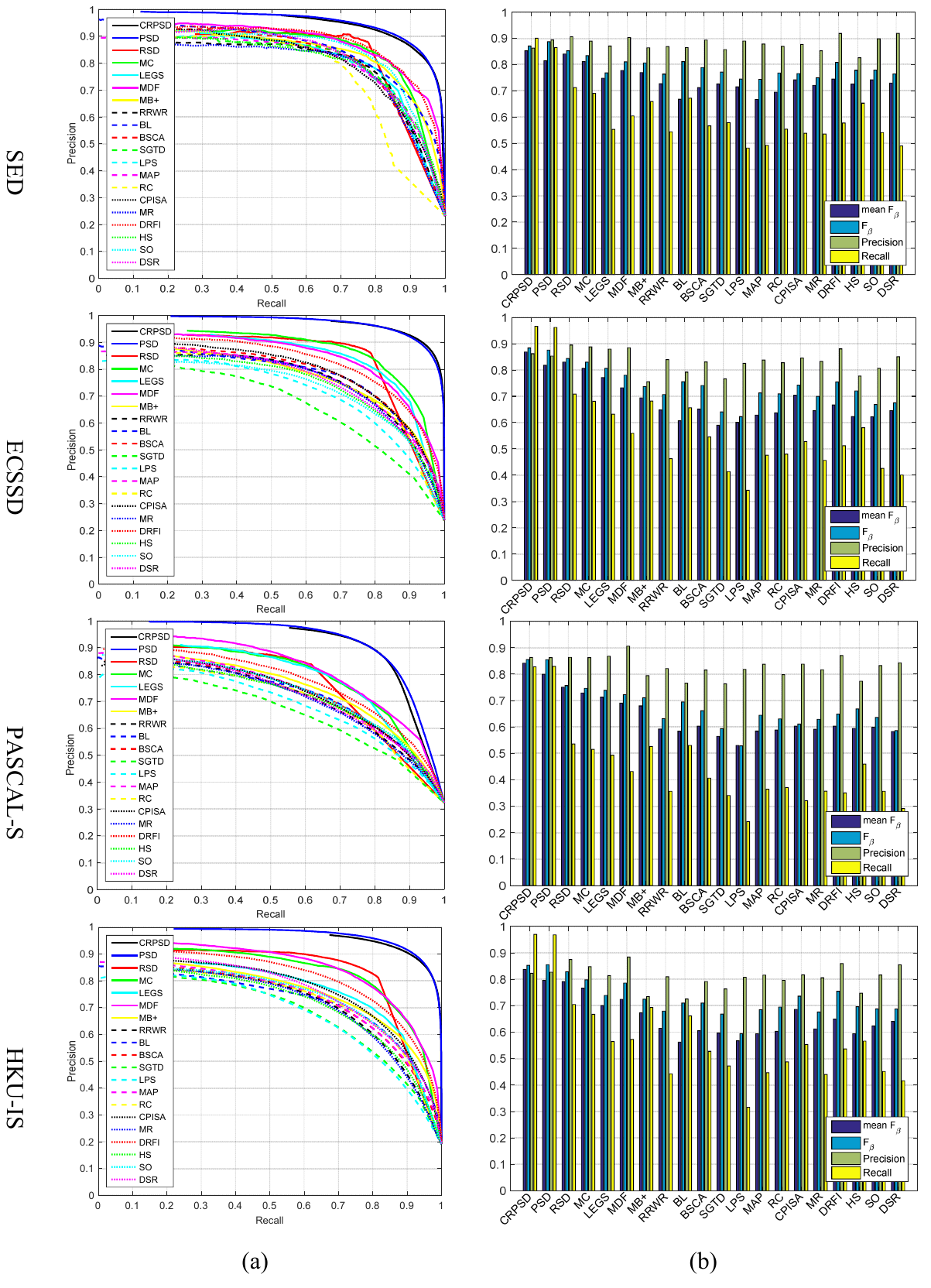}
	\caption{Results of all test approaches on four standard benchmark datasets, i.e. SED, ECSSD, PASCAL-S, and HKU-IS. (a) presents the PR curves, (b) presents the mean $F_\beta$ and the adaptive $F_\beta$/precision/recall which are computed from the binary images obtained by using Otsu algorithm on the salient maps.}
	\label{fig8}
\end{figure}

According to Fig. \ref{fig8} and Table \ref{tbl1}, the proposed method (CRPSD) significantly outperforms all of the state-of-the-art approaches on all test datasets in terms of all evaluation criterions, which convincingly demonstrates the effectiveness of the proposed method. In these four test datasets, the most complex one is PASCAL-S. Therefore, all methods get the worst performance on this dataset. For all datasets, our method gets the largest gain on PASCAL-S dataset compared with the best state-of-the-art approach (MC) or our PSD, which demonstrates that our method can better deal with the complex cases than other approaches. 

From the experimental results, three benefits of our method can be confirmed. (1) Although only the submodule region-level saliency estimation is used, it still gets the best performance compared with the state-of-the-art approaches on four datasets. Compared with MC \cite{26zhao2015saliency}, the RSD estimates the region saliency based on the regions generated by the proposed adaptive region generation technique while MC is based on superpixels, and the RSD uses a different strategy to form the context windows. The good performance of the RSD demonstrates the effectiveness of these improvements. (2) The submodule PSD also gets the best performance compared with the state-of-the-art approaches, which validates that the pixel-level CNN modified from VGGNet can well extract the multi-scale deep features for pixels to decide its saliency. (3) The proposed CRPSD by using the fusion network and joint training with the pixel-level CNN network can greatly improve the performance of the submodules, which demonstrates that CRPSD can well dig the complementary information of saliencies estimated by RSD and PSD for saliency detection.

\setlength{\tabcolsep}{2.7pt}
\begin{table}[!htb]
	\begin{center}
		\caption{The $wF_\beta$ and $ MAE $ of different saliency detection method on different test datasets (\textcolor{red}{red}, \textcolor{blue}{blue}, and \textcolor{green}{green} texts respectively indicate rank 1, 2, and 3).}
		\label{tbl1}
		\begin{tabular}{cccccccccc}
			\hline\noalign{\smallskip}
			\multirow{2}{*}{Method} & \multirow{2}{*}{Year} & \multicolumn{2}{c}{SED} & \multicolumn{2}{c}{ECSSD} & \multicolumn{2}{c}{PASCAL-S} & \multicolumn{2}{c}{HKU-IS}\\
			\cline{3-10}
			& & $wF_\beta$ & $MAE$ & $wF_\beta$ & $MAE$ & $wF_\beta$ & $MAE$ & $wF_\beta$ & $MAE$ \\
			\noalign{\smallskip}
			\hline
			\noalign{\smallskip}
			\textbf{CRPSD} & / & \textcolor{red}{0.8292} & \textcolor{red}{0.0509} & \textcolor{red}{0.8485} & \textcolor{red}{0.0455} & \textcolor{red}{0.7761} & \textcolor{red}{0.0636} &	\textcolor{red}{0.8209} & \textcolor{red}{0.0431} \\
			\textbf{PSD} & / & \textcolor{green}{0.7590} & \textcolor{blue}{0.0758} & \textcolor{blue}{0.7572} & \textcolor{blue}{0.0798} & \textcolor{blue}{0.7113} & \textcolor{blue}{0.1057} & \textcolor{blue}{0.7371} & \textcolor{blue}{0.0693} \\
			\textbf{RSD} & / & \textcolor{blue}{0.7759} & \textcolor{green}{0.0922} & \textcolor{green}{0.7569} & \textcolor{green}{0.0915} & \textcolor{green}{0.6195} & \textcolor{green}{0.1338} & \textcolor{green}{0.7286} & \textcolor{green}{0.0813} \\
			MC & CVPR2015 & 0.7387 & 0.1032 & 0.7293 & 0.1019 & 0.6064 & 0.1422 & 0.6899 & 0.0914 \\
			LEGS & CVPR2015 & 0.6498 & 0.1279 & 0.6722 & 0.1256 & 0.5791 & 0.1593 & 0.5911 & 0.1301 \\
			MDF & CVPR2015 & 0.6748 & 0.1196 & 0.6194 & 0.1377 & 0.5386 & 0.1633 & 0.6135 & 0.1152 \\
			MB+ & ICCV2015 & 0.6555 & 0.1364 & 0.5632 & 0.1717 & 0.5307 & 0.1964 & 0.5438 & 0.1497 \\
			RRWR & CVPR2015 & 0.6117 & 0.1547 & 0.5026 & 0.1850 & 0.4435 & 0.2262 & 0.4592 & 0.1719 \\
			BL & CVPR2015 & 0.4986 & 0.1887 & 0.4615 & 0.2178 & 0.4464 & 0.2478 & 0.4119 & 0.2136 \\
			BSCA & CVPR2015 & 0.5671 & 0.1576 & 0.5159 & 0.1832 & 0.4703 & 0.2220 & 0.4643 & 0.1760 \\
			SGTD & TIP2015 & 0.6216 & 0.1475 & 0.4689 & 0.2007 & 0.4385 & 0.2269 & 0.4785 & 0.1627 \\
			LPS & TIP2015 & 0.5976 & 0.1477 & 0.4585 & 0.1877 & 0.3882 & 0.2162 & 0.4252 & 0.1635 \\
			MAP & TIP2015 & 0.5567 & 0.1621 & 0.4953 & 0.1861 & 0.4361 & 0.2222 & 0.4533 & 0.1717 \\
			RC & TPAMI2015 & 0.5652 & 0.1588 & 0.5118 & 0.1868 & 0.4694 & 0.2253 & 0.4768 & 0.1714 \\
			CPISA & TIP2015 & 0.6174 & 0.1474 & 0.5735 & 0.1596 & 0.4478 & 0.1983 & 0.5575 & 0.1374 \\
			MR & CVPR2013 & 0.6052 & 0.1586 & 0.4985 & 0.1875 & 0.4406 & 0.2288 & 0.4556 & 0.1740 \\
			DRFI & CVPR2013 & 0.6464 & 0.1360 & 0.5433 & 0.1658 & 0.4817 & 0.2042 & 0.5180 & 0.1444 \\
			HS & CVPR2013 & 0.5828 & 0.1948 & 0.4571 & 0.2283 & 0.4516 & 0.2625 & 0.4213 & 0.2151 \\
			SO & CVPR2014 & 0.6568 & 0.1351 & 0.5134 & 0.1733 & 0.4723 & 0.1986 & 0.5162 & 0.1426 \\
			DSR & ICCV2013 & 0.6055 & 0.1476 & 0.5162 & 0.1728 & 0.4385 & 0.2043 & 0.5079 & 0.1429 \\
			\hline
		\end{tabular}
	\end{center}
\end{table}

\begin{figure}[!htb]
	\centering
	\includegraphics[width=0.99\textwidth]{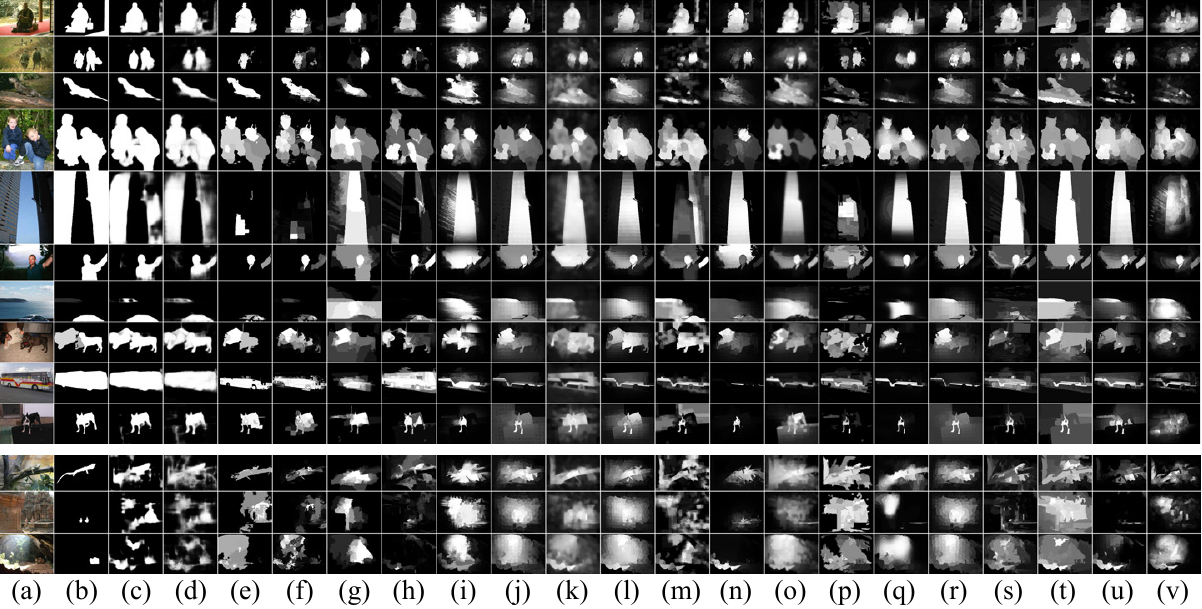}
	\caption{Visual Comparisons of different saliency detection approaches in various challenging scenarios. (a) Original images, (b) Ground truths, (c) CRPSD, (d) PSD, (e) RSD, (f) MC, (g) LEGS, (h)MDF, (i) MB+, (j) RRWR, (k) BL, (l) BSCA, (m) SGTD, (n) LPS, (o) MAP, (p) RC, (q) CPISA, (r) MR, (s) DRFI, (t) HS, (u) SO, (v) DSR.}
	\label{fig9}
\end{figure}

Also, we qualitatively compare the salient maps detected by different approaches, as shown in the first ten rows of Fig. \ref{fig9}. Obviously, the proposed method is able to highlight saliencies of salient objects and suppress the saliencies of background better than other approaches, and the salient maps of the proposed method are much close to the ground truths in various challenging scenarios. 

The last three rows of Fig. \ref{fig9} show some cases in which the proposed method fails. For example, the colors of salient objects and backgrounds are very similar, the salient objects are too small, and the backgrounds are too complex. In these cases, the other approaches also cannot correctly detect the salient objects and it is not easy to accurately locate the salient objects even for human eyes.

\subsection{Performance comparisons with baselines}
As pixel labeling task, saliency detection and semantic segmentation are very similar. And recently, many CNN models \cite{38long2015fully,37zheng2015conditional,chen2016deeplab} have been proposed for semantic segmentation. In order to test their performance on saliency detection, the most powerful model of deeplab \cite{chen2016deeplab}, i.e. the DeepLab-MSc-LargeFOV model (DML), is chosen as a baseline, which is trained on THUS-10000 dataset for saliency detection. And its pretrained DeepLab-LargeFOV-COCO-MSC model (pre-DML) on semantic image segmentation is used as another baseline, which is directly used for saliency detection by summing up the probability predictions across all 20 object classes and using these sumed-up probabilities as a salient map. And to demonstrate the benefit of joint training of our method, we also test the performance of our method with separate training (sep-CRPSD).

\setlength{\tabcolsep}{8pt}
\begin{table}[!htb]
	\begin{center}
		\caption{The $wF_\beta$ of baselines and our methods on all test datasets.}
		\label{tbl2}
		\begin{tabular}{ccccc}
			\hline\noalign{\smallskip}
			Method & SED & ECSSD & PASCAL-S & HKU-IS \\
			\noalign{\smallskip}
			\hline
			\noalign{\smallskip}
			pre-DML & 0.5140 & 0.6530 & 0.7322 & 0.6755 \\
			DML & 0.7439 & 0.7482 & 0.6948 & 0.7258 \\
			sep-CRPSD & 0.8109 & 0.8249 & 0.7621 & 0.7942 \\
			\textbf{CRPSD} & \textbf{0.8292} & \textbf{0.8485} & \textbf{0.7761} & \textbf{0.8209} \\
			\hline
		\end{tabular}
	\end{center}
\end{table}

Table \ref{tbl2} lists the $wF_\beta$ of baselines and our methods on all test datasets. According to Table \ref{tbl2}, three conclusions can be summarized: (1) The performance of pre-DML is very good on PASCAL-S, while dramatically drops on other datasets. Because many salient objects in other datasets don't belong to the trained classes, and hence are considered as non-salient objects during saliency detection. (2) The DML trained for saliency detection gets better results than pre-DML on all datasets except PASCAL-S, but still much worse than our method, which further demonstrates that our method with multiple CNNs is powerful for saliency detection. (3) Our method with joint training (CRPSD) gets better performance than separate training (sep-CRPSD), which demonstrates the effectiveness of joint training.

\subsection{Performance of fixation prediction with pixel-level CNN}
The model (PSD) for pixel-level saliency prediction also can be used for fixation prediction. To validate its performance for fixation prediction, we use the same experimental setting with Mr-CNN \cite{liu2015predicting} to test our model on MIT \cite{judd2009learning} and Toronto \cite{bruce2009saliency} datasets. The evaluation metric is mean shuffled-AUC \cite{zhang2016exploiting}. Table \ref{tbl3} lists the experimental results of our model and the other three state-of-the-art fixation prediction approaches on these two datasets. According to Table \ref{tbl3}, PSD gets the best performance, which means that our model has powerful ability of fixation prediction. Above experimental results further demonstrate the effectiveness of our pixel-level CNN model.

\setlength{\tabcolsep}{8pt}
\begin{table}[!htb]
	\begin{center}
		\caption{The mean shuffled-AUC of different fixation prediection methods on test datasets.}
		\label{tbl3}
		\begin{tabular}{ccccc}
			\hline\noalign{\smallskip}
			Dataset & \textbf{PSD} & Mr-CNN \cite{liu2015predicting} & SDAE \cite{han2016two} & BMS \cite{zhang2016exploiting} \\
			\noalign{\smallskip}
			\hline
			\noalign{\smallskip}
			MIT & \textbf{0.7587} & 0.7184 & 0.7095 & 0.7105 \\
			Toronto & \textbf{0.7606} & 0.7221 & 0.7230 & 0.7243 \\
			\hline
		\end{tabular}
	\end{center}
\end{table}

\section{Conclusions}
This paper proposes a novel saliency detection method by combining region-level saliency estimation and pixel-level saliency prediction (denoted as CRPSD). A multiple CNN framework, composed of pixel-level CNN, region-level CNN and fusion CNN, is proposed for saliency detection. The pixel-level CNN, which is a modification of VGGNet, can predict the saliency at pixel-level by extracting multi-scale features of images. The region-level CNN can effectively estimate the saliencies of these regions generated by the proposed adaptive region generation technique. The fusion CNN can take full advantage of the original image, the pixel-level and region-level saliencies for final saliency detection. The proposed method can effectively detect the salient maps of images in various scenarios and greatly outperform the state-of-the-art saliency detection approaches.

\section*{Acknowledgements}
This work was supported by the Natural Science Foundation of China under Grant 61472102. The authors would like to thank the founders of the publicly available datasets and the support of NVIDIA Corporation with the donation of the Tesla K40 GPU used for this research.

\bibliographystyle{splncs}
\bibliography{egbib}
\end{document}